\documentclass{article}

\usepackage{arxiv}

\usepackage[utf8]{inputenc} % allow utf-8 input
\usepackage[T1]{fontenc}    % use 8-bit T1 fonts
\usepackage{hyperref}       % hyperlinks
\usepackage{url}            % simple URL typesetting
\usepackage{graphicx}
\usepackage{subcaption}
\usepackage{booktabs}       % professional-quality tables
\usepackage{amsfonts}       % blackboard math symbols
\usepackage{nicefrac}       % compact symbols for 1/2, etc.
\usepackage{microtype}      % microtypography
\usepackage{lipsum}
\usepackage{graphicx}
\usepackage{pifont}
\graphicspath{ {./images/} }
\usepackage{amsmath}

\title{CogniSQL-R1-Zero: Lightweight Reinforced Reasoning for Efficient SQL Generation}

\author{
 Kushal Gajjar \\
  Dell Technologies\\
  \texttt{kushalgajjar1@gmail.com} \\
   \And
 Harshit Sikchi \\
  UT Austin\\
  \texttt{hsikchi@utexas.edu} \\
  \And
 Arpit Singh Gautam \\
  Dell Technologies\\
  \texttt{arpitsinghgautam777@gmail.com} \\
  \And
Marc Hammons \\
  Dell Technologies\\
  \texttt{marchammons@gmail.com} \\
  \And
Saurabh Jha \\
  Dell Technologies\\
  \texttt{saurabh.jha21@gmail.com} \\
}

\begin{document}
\maketitle
\begin{abstract}
Translating natural language into SQL (Text-to-SQL) remains a core challenge at the intersection of language understanding and structured data access. Although large language models (LLMs) have improved fluency, generating correct and executable SQL, especially for complex queries, continues to be challenging. We introduce \textbf{CogniSQL-R1-Zero}, a reinforcement learning (RL) framework and model that produces accurate SQL using a lightweight reward signal based on execution correctness and format‐tag compliance. By avoiding intermediate supervision, hybrid pipelines and complex reward shaping, our method encourages stable learning and stronger alignment with the ultimate task objective—producing executable programs. CogniSQL-R1-Zero achieves state-of-the-art execution accuracy on  Text2SQL benchmark; BIRD bench, outperforming prior supervised and instruction-tuned baselines including SFT CodeS-7B, DeepSeek-Coder 236B, and Mistral 123B—despite being trained on a significantly \textit{smaller 7B backbone}. This result underscores the scalability and efficiency of our RL-based approach when trained on just four NVIDIA A100 GPUs (40 GB VRAM each). To support further research in efficient and interpretable Text-to-SQL modeling, we release two curated datasets: (i) a collection of 5,024 reasoning traces with varying context lengths, and (ii) a positive-sampled corpus of 36,356 corpus of weakly supervised queries, each annotated with six semantically diverse reasoning paths. Together, these contributions advance scalable, execution-aligned Text-to-SQL generation. \\

\url{https://huggingface.co/datasets/CogniSQL/Reasoning_Traces} \\
\url{https://huggingface.co/datasets/CogniSQL/Positive_Sample_Corpus}
\end{abstract}

% keywords can be removed
%\keywords{First keyword \and Second keyword \and More}

% \section{Introduction}

\section{Introduction}
\label{sec:introduction}

Translating natural language questions into SQL queries—commonly referred to as Text-to-SQL—has long been a cornerstone problem at the intersection of natural language understanding and structured data retrieval. Despite substantial improvements in fluency and token-level correctness provided by large language models (LLMs), ensuring that generated SQL is both syntactically valid and semantically executable remains a persistent challenge. The complexity arises from (i) the inherent ambiguity and variability of human language, (ii) heterogeneous and intricate database schemas, and (iii) the gap between surface-level fluency and true execution correctness. As a result, even state-of-the-art LLMs often produce queries that “look” plausible but fail when run against a real database.

In investigations leading to this work, we first experimented with hybrid pipelines that leveraged LLaMA 3.1 8B Instruct for reasoning and CodeStral 22B for SQL generation. We also developed an agentic framework with agents running in parallel to handle complex business contexts, finally feeding consolidated reasoning into a Qwen-based SQL generator. Using these approaches - combined with techniques such as chain-of-thought prompting \cite{wei2023chain_of_thought}, query decomposition, and contextual harnessing \cite{talaei2024chess} - we achieved up to 85\% execution accuracy on 200 random sample queries from the data set. However, despite extensive tuning and ensemble methods (e.g., XiYan-SQL \cite{gao2024xiyan_sql} and DIN-SQL \cite{guo2019towards}), we were unable to surpass the mark on complex, real-world queries. This gap motivated a shift toward explicit reinforcement learning (RL)–based training, inspired by recent work such as DeepSeek-R1 \cite{guo2025deepseek_r1} and related RL-driven Text-to-SQL efforts \cite{pourreza2025reasoning_sql, chen2025graph_reward_sql}.

In this paper, we introduce \textbf{CogniSQL-R1-Zero}, an RL-trained Text-to-SQL framework that optimizes directly for execution correctness rather than relying on brittle intermediate labels or complex reward shaping. By adopting Group Relativle Policy Optimization (GRPO) \cite{zhong2017seq2sql} and a structured prompt format—consisting of DDL, external knowledge, natural language questions, and explicit format instructions—CogniSQL-R1-Zero aligns model behavior tightly with the end task. Trained on four NVIDIA A100 GPUs (40GB each) using DeepSpeed ZeRO 2 \cite{rasley2020deepspeed} and gradient accumulation, our 7B-parameter model achieves \textbf{59.97\% execution accuracy on the BIRD dev set}, outperforming much larger baselines (e.g., GPT-4, Mistral 123B, DeepSeek-Coder 236B) as well as 7B/8B models and SFT CodeS-7B.

Our primary contributions are as follows:
\begin{itemize}
    \item \textbf{State-of-the-Art Benchmark Results:} CogniSQL-R1-Zero achieves 59.97\% execution accuracy on the BIRD development set—surpassing larger models (GPT-4, Mistral 123B, DeepSeek-Coder 236B) and 7B/8B baselines, including SFT CodeS-7B—demonstrating the efficacy of RL training under low-compute constraints (4 NVIDIA A100 GPUs having 40 GB VRAM each).
    \item \textbf{Open-Source Datasets for Reasoned SQL:} We release two curated datasets to support future work: (i) a \textit{positive-sampling corpus} of approximately 36,356 examples (correct samples from six semantically diverse reasoning paths per query) generated by Qwen-7B-Coder at temperature 0.9, and (ii) \textit{QWQ 32B reasoning traces} comprising around 5,024 examples with context lengths varying across queries. These resources enable alignment-driven RL training for any base LLM.
    \item \textbf{Practical Inference-Time Techniques:} We show that lightweight strategies—such as retrieving schema-specific values during query generation and aggregating multiple candidates via majority voting—can boost execution accuracy with negligible overhead, underscoring CogniSQL-R1-Zero’s readiness for real-world deployment.
    \item \textbf{Data and Training Strategies That Matter:} Through extensive ablation studies, we quantify the impact of prompt design, sample curation, and RL versus SFT initialization, offering actionable guidelines for efficient Text-to-SQL development in resource-constrained environments.
    \item \textbf{Empirical Guidance:} We present a comprehensive set of observations—highlighting both successful strategies and pitfalls—to equip the community with practical takeaways for RL-driven Text-to-SQL development.
\end{itemize}

The remainder of this paper is organized as follows. Section~\ref{sec:background} surveys related Text-to-SQL and RL-based methods. Section~\ref{sec:methodology} outlines our structured prompt design, reward formulation, and GRPO training. Section~\ref{sec:learnings} details our experimental setup, low-compute optimizations, and dataset curation. Section~\ref{sec:results} presents empirical findings and ablation analyses. Section~\ref{sec:discussion} discusses limitations, insights, and future directions. Finally, Section~\ref{sec:conclusion} concludes with a summary and avenues for future work.

\section{Background}
\label{sec:background}

Translating natural language questions into SQL queries (Text-to-SQL) dates back decades as a key interface between unstructured inputs and structured data stores~\cite{androutsopoulos1995natural}. Early rule-based systems relied on handcrafted grammars and template matching, but struggled to generalize across domains. The advent of neural sequence-to-sequence models marked a major shift, enabling data-driven learning of mappings from questions to SQL.

\subsection{Neural Text-to-SQL and Seq2SQL}
% One of the pioneering neural approaches, Seq2SQL~\cite{zhong2017seq2sql}, decomposes SQL generation into sub-tasks (aggregation, selection, conditions) and introduces reinforcement learning for the unordered WHERE clause. Its combined objective is:
% \begin{equation}
% \mathcal{L} = \mathcal{L}_{\text{CE}} + \lambda \,\mathbb{E}_{q \sim \pi_\theta}\bigl[R(q)\bigr],
% \end{equation}
% where $\mathcal{L}_{\text{CE}}$ is the cross-entropy loss over supervised SQL tokens, $\pi_\theta$ the policy for sampling queries $q$, and $R(q)$ an execution-based reward. This strategy improved execution accuracy by aligning training with query execution outcomes.

One of the pioneering neural approaches, Seq2SQL~\cite{zhong2017seq2sql}, decomposes SQL generation into sub-tasks—aggregation, selection, and condition construction—and employs reinforcement learning for the unordered WHERE clause. The objective combines supervised learning for aggregation and selection heads with a policy-gradient term for the WHERE clause:

\begin{equation}
\mathcal{L}_{\mathrm{Seq2SQL}} = \mathcal{L}_{\mathrm{agg}} + \mathcal{L}_{\mathrm{sel}} + \mathcal{L}_{\mathrm{whe}},
\end{equation}

where the cross-entropy terms \(\mathcal{L}_{\mathrm{agg}}\) and \(\mathcal{L}_{\mathrm{sel}}\) supervise aggregation and column selection heads, respectively. The reinforcement component is:

\begin{equation}
\mathcal{L}_{\mathrm{whe}} = -\mathbb{E}_{q\sim\pi_\theta}\left[ R(q)\sum_{t=1}^{T_{\mathrm{whe}}} \log \pi_{\theta}(q_t \mid q_{<t})\right],
\end{equation}

and the execution-based reward \(R(q)\) assigns:
\[
R(q) = \begin{cases}
+1, & \text{if valid SQL and correct result},\\
-1, & \text{if valid SQL but incorrect result},\\
-2, & \text{if invalid SQL},
\end{cases}
\]

based on execution outcomes~\cite{zhong2017seq2sql}. This integration aligns training with actual execution, improving end-to-end performance on datasets like WikiSQL by encouraging the model to generate executable, semantically correct WHERE clauses.

\subsection{Advances in Large Language Models}
Recent large language models (LLMs) such as GPT-3/4 and open-source variants have demonstrated strong few-shot Text-to-SQL capabilities~\cite{gao2023text2sql_llm}. Structural enhancements—e.g., UniSAr’s schema-aware markers~\cite{dou2022unisar}—and constrained decoding mechanisms enforce SQL syntax during generation. For instance, the probability of generating a SQL token sequence $S = (s_1,\dots,s_T)$ given question $Q$ and schema context $C$ is
\begin{equation}
P(S \mid Q, C) \;=\; \prod_{t=1}^{T} P\bigl(s_t \mid s_{<t},\,Q,\,C\bigr).
\end{equation}
However, LLMs often produce syntactically plausible but semantically incorrect queries, motivating methods that incorporate execution feedback.

% \subsection{Hybrid and Agentic Pipelines}
% To leverage reasoning capabilities, we experimented with hybrid pipelines that separate reasoning from SQL synthesis. Initially, we used \textbf{LLaMA 3.1 8B} to perform multi-step reasoning (chain-of-thought~\cite{wei2023chain_of_thought}) and \textbf{CodeStral 22B} to generate SQL. In parallel, an \emph{agentic} framework spawned four specialized reasoning agents—each maintaining context for different business scenarios—before combining outputs into a final Qwen-based SQL generator. This hybrid approach, augmented by CHESS-style contextual harnessing~\cite{talaei2024chess}, achieved \textbf{85\% execution accuracy} on a proprietary Dell dataset, but failed to surpass 90\%. These findings highlighted the remaining gap between surface fluency and reliable execution correctness.

\subsection{Hybrid and Agentic Pipelines}
To leverage chain-of-thought reasoning capabilities, we first experimented with a \textbf{hybrid chain-of-thought decomposition pipeline}. In this setup, \textbf{LLaMA 3.1 8B} served as a reasoning engine that decomposed a complex question \(Q\) into a sequence of intermediate reasoning states:
\begin{align}
r^{(0)} &= \text{encode}(Q), \\
r^{(k)} &= f_{\text{reason}}\bigl(r^{(k-1)},\,Q\bigr), \quad k = 1, \dots, K,
\end{align}
where \(r^{(K)}\) is the final reasoning representation after \(K\) steps. The SQL generator (\textbf{CodeStral 22B}) then produced a query \(S\) conditioned on the reasoning trace:
\begin{equation}
P(S \mid Q) 
= P\bigl(S \mid r^{(K)}\bigr) 
= \prod_{t=1}^{T} P\bigl(s_t \mid s_{<t},\,r^{(K)}\bigr).
\end{equation}
This two-stage process—reasoning followed by generation—yielded approximately \(\mathbf{75\%}\) execution accuracy on 200 random sample queries from the data set. However, it required two full LLM invocations per query, leading to high inference latency and resource consumption.

Next, we built an \textbf{agentic reasoning pipeline}, in which four specialized agents (\(a = 1,2,3,4\)) ran in parallel, each focusing on a distinct business context. By reducing token sizes and parallelizing inference calls, this framework significantly lowered latency compared to the hybrid pipeline while still achieving the \(\mathbf{desired\;85\%}\) execution accuracy on the internal dataset. Each agent produced its own reasoning trace \(r_a^{(K)}\) and candidate SQL \(S_a\):
\begin{align}
r_a^{(0)} &= \text{encode}_a(Q), \\
r_a^{(k)} &= f_{\text{reason},a}\bigl(r_a^{(k-1)},\,Q\bigr), \quad k = 1, \dots, K, \\
P\bigl(S_a \mid Q\bigr) 
&= \prod_{t=1}^{T} P\bigl(s_{a,t} \mid s_{a,<t},\,r_a^{(K)}\bigr).
\end{align}
These candidate queries were then aggregated via majority voting:
\begin{equation}
S^* = \operatorname{arg\,max}_{S} \Bigl\lvert \{\,a : S_a = S\}\Bigr\rvert,
\end{equation}
where \(S^*\) is the final selected query. Although parallelism reduced inference time, the cost of running multiple large models simultaneously remained high, and the system complexity and memory footprint proved impractical for broader deployment.

Despite these gains, neither approach was cost-effective or sufficiently low-latency. Motivated by the release of DeepSeek-R1~\cite{guo2025deepseek_r1} and related RL-based Text-to-SQL work~\cite{pourreza2025reasoning_sql, chen2025graph_reward_sql}, we shifted to an \textbf{RL-trained reasoning paradigm}, condensing reasoning and SQL generation into a single GRPO-trained model to reduce both latency and cost while directly optimizing for execution correctness.

% \subsection{Reinforcement Learning for Reasoned SQL}
% Inspired by DeepSeek-R1~\cite{guo2025deepseek_r1} and related RL approaches~\cite{pourreza2025reasoning_sql, chen2025graph_reward_sql}, we shifted to training models directly with an execution-based reward. Group Relative Policy Optimization (GRPO)~\cite{zhong2017seq2sql} stabilizes policy updates by grouping similar queries. Denoting $\Theta$ as model parameters and $J(\theta) = \mathbb{E}_{q \sim \pi_\theta}[R(q)]$ as the expected reward, GRPO updates optimize:
% \begin{equation}
% \nabla_\theta J(\theta) \;\approx\; \frac{1}{N}\sum_{i=1}^N \bigl(R(q_i) - b\bigr)\,\nabla_\theta \log \pi_\theta(q_i),
% \end{equation}
% where $b$ is a baseline to reduce variance. This direct reinforcement objective encourages generation of executable SQL without intermediate supervision.

% \subsection{Motivation for Lightweight Reasoning}
% Our internal experiments underscored that large, multi-stage pipelines impose high latency and resource costs, and often rely on brittle annotations. By contrast, an RL-trained model using only execution feedback can (i) simplify training, (ii) align directly with the end task, and (iii) run effectively on limited hardware. We therefore focus on designing \emph{CogniSQL-R1-Zero} as a \textbf{lightweight} RL-trained reasoning model in order to democratize reliable Text-to-SQL for practical, low-compute environments.

\subsection{Reinforcement Learning for Reasoned SQL}
\label{sec:rl-reasoned-sql}

Recent advancements in reinforcement learning (RL) have significantly enhanced the reasoning capabilities of large language models (LLMs) in Text-to-SQL generation. Traditional supervised methods rely on labeled SQL pairs, which can be expensive to curate and may not capture execution semantics. In contrast, RL enables models to learn directly from execution feedback, using a reward signal based solely on whether the generated SQL returns correct results.

% One effective RL technique is \textbf{Group Relative Policy Optimization (GRPO)}~\cite{zhong2017seq2sql}, which optimizes over groups of candidate queries rather than individual tokens. Let $\pi_\theta$ denote the model’s policy parameterized by $\theta$. Given a natural language question $Q$, the model samples a set of $N$ SQL candidates $\{q_i\}_{i=1}^N$. Each candidate $q_i$ receives a binary execution reward:
% \[
% R(q_i) = 
% \begin{cases}
% 1, & \text{if } \mathrm{Exec}(q_i) = \mathrm{Exec}(q_{\text{gt}}) \\
% 0, & \text{otherwise}
% \end{cases}
% \]
% where $\mathrm{Exec}(\cdot)$ returns the execution result on the database, and $q_{\text{gt}}$ is the ground-truth SQL. The expected group reward is:
% \begin{equation}
% J(\theta) \;=\; \mathbb{E}_{\{q_i\}\sim \pi_\theta}\Bigl[\max_{i} R(q_i)\Bigr].
% \end{equation}
% GRPO updates $\theta$ by comparing groups of candidates, improving stability:
% \begin{equation}
% \nabla_\theta J(\theta)
% \;\approx\;
% \frac{1}{M} \sum_{m=1}^{M} \Bigl(\max_{i} R(q_i^{(m)}) - b\Bigr)\, \nabla_\theta \log \pi_\theta \bigl(q_{i^*}^{(m)}\bigr),
% \end{equation}
% where $b$ is a baseline (e.g., average reward) and $q_{i^*}^{(m)}$ is the top-reward candidate in group $m$.

One effective RL technique is \textbf{Group Relative Policy Optimization (GRPO)}, introduced in DeepSeek-R1~\cite{guo2025deepseek_r1}. GRPO optimizes over groups of candidate SQL queries rather than individual token-level gradients. Let \(\pi_\theta\) denote the model’s policy parameterized by \(\theta\). For each question \(Q\), the model samples a set of \(N\) SQL candidates \(\{q_i\}_{i=1}^N\), each receiving a binary execution-based reward:
\[
R(q_i) = 
\begin{cases}
1, & \text{if } \mathrm{Exec}(q_i) = \mathrm{Exec}(q_{\text{gt}}), \\
0, & \text{otherwise},
\end{cases}
\]
where \(\mathrm{Exec}(\cdot)\) returns the execution result, and \(q_{\text{gt}}\) is the ground-truth SQL. The expected group reward becomes:
\[
J(\theta) = \mathbb{E}_{\{q_i\}\sim \pi_\theta}\Bigl[\max_i R(q_i)\Bigr].
\]
Policy updates are performed over groups to improve training stability:
\[
\nabla_\theta J(\theta)
\approx
\frac{1}{M} \sum_{m=1}^{M} \Bigl(\max_i R(q_i^{(m)}) - b^{(m)}\Bigr) \, \nabla_\theta \log \pi_\theta\bigl(q_{i^*}^{(m)}\bigr),
\]
where \(b^{(m)}\) is the group’s baseline (e.g., mean reward), and \(q_{i^*}^{(m)}\) is the top-reward candidate in group \(m\). This update is augmented by a PPO-style clipped surrogate objective and KL-divergence penalty to encourage stable policy evolution while eliminating the need for a separate value critic~\cite{guo2025deepseek_r1}. The group-based advantage normalization reduces variance and enables efficient optimization in large language models.

Another RL approach uses \textbf{reverse curriculum learning} to ease the reasoning difficulty gradually. Starting from simpler SQL templates, the model learns to solve easier queries before tackling complex multi-join statements. Denote the curriculum level by $c$, where $c=0$ corresponds to single-table queries and $c=K$ to full multi-join queries. At each level:
\begin{align}
r^{(0)} &= \text{encode}(Q), \\
r^{(k)} &= f_{\text{reason}}^{(c)}\bigl(r^{(k-1)}, Q\bigr), \quad k=1,\dots,K_c,
\end{align}
with $K_c$ decreasing as $c$ increases. The policy $\pi_\theta^{(c)}$ is trained at each level $c$ with its own reward $R_c(q)$, then the model is smoothly transitioned to $c+1$ once a performance threshold is met.

These RL-based methods, exemplified by DeepSeek-R1~\cite{guo2025deepseek_r1} and Graph-Reward-SQL~\cite{chen2025graph_reward_sql}, demonstrate that LLMs can acquire robust reasoning without detailed intermediate annotations, relying instead on execution feedback to guide learning toward accurate, executable SQL.

\subsection{Motivation for Lightweight Reasoning}
\label{sec:motivation-lightweight}

Deploying LLM-based Text-to-SQL systems in real-world environments often confronts two critical constraints: \textbf{computational cost} and \textbf{inference latency}. Large, multi-stage reasoning pipelines may yield high accuracy but are impractical for low-resource deployment. Consequently, lightweight reasoning models—which preserve performance while reducing compute and latency—are gaining attention.

One effective strategy is \textbf{dynamic resource allocation}, where computation adapts to the query complexity. Let \(C(Q)\) measure the complexity (e.g., number of tables or joins). The compute budget is allocated as:

\[
B(Q) = B_{\min} + \alpha \cdot C(Q),
\]

where \(B_{\min}\) ensures coverage for simple queries, and \(\alpha\) scales compute for harder ones. This allows skipping redundant layers or reducing token processing, improving throughput without sacrificing accuracy.

A complementary method is \textbf{information-theoretic reward shaping}, inspired by the \textbf{Learning to Think (L2T)} framework~\cite{zhang2025l2t}. For each reasoning step \(k\), the information gain is computed as:

\[
\Delta I_k = H(r^{(k-1)}) - H\bigl(r^{(k)}\bigr),
\]

where \(H(\cdot)\) is the entropy of the reasoning state. The RL objective maximizes:

\[
J(\theta) = \mathbb{E}\Bigl[\sum_{k=1}^{K} \gamma^{k-1} \Delta I_k - \beta \cdot \ell_{\text{len}}\bigl(r^{(k)}\bigr)\Bigr],
\]

with discount \(\gamma\) and length penalty \(\ell_{\text{len}}\). This training encourages concise, informative reasoning chains.

Finally, \textbf{policy distillation} transfers capabilities from a heavy RL-tuned teacher \(\pi_T\) to a smaller student \(\pi_S\):

\[
\mathcal{L}_{\text{distill}} = 
\mathrm{KL}\bigl(\pi_T(\cdot\mid Q)\,\|\,\pi_S(\cdot\mid Q)\bigr)
+ \lambda \cdot \mathbb{E}_{q\sim\pi_S}\bigl[1 - R(q)\bigr],
\]

where \(R(q)\) is the execution-based reward. Prior work shows this enables significantly smaller models to maintain performance.

These insights directly inform \textbf{CogniSQL-R1-Zero}: a unified, lightweight reasoning model designed for efficient Text-to-SQL under compute constraints.

\section{Related Work}
\label{sec:related}

The research presented in CogniSQL-R1-Zero advances two key domains: (i) reinforcement learning–driven reasoning in large language models (LLMs), and (ii) Text-to-SQL methodologies, including dataset curation and reward engineering.

\subsection{Reinforcement Learning for LLM Reasoning}
Recent studies have shown that reinforcement learning (RL) can substantially improve LLM reasoning without requiring dense supervision. For example, \textbf{DeepSeek-R1} trains LLMs purely via execution-based reward signals, demonstrating emergent multi-step reasoning and self-verification capabilities~\cite{guo2025deepseek_r1}. Similarly, \textbf{Graph-Reward-SQL} uses graph-based representations to provide reward feedback without executing queries, reducing inference cost while maintaining reasoning quality~\cite{chen2025graph_reward_sql}.

\textbf{Group Relative Policy Optimization (GRPO)} is a pivotal algorithm in this space, as it evaluates groups of candidate outputs rather than individual tokens~\cite{zhong2017seq2sql, li2025grpo}. Let $\pi_\theta$ denote the model policy over SQL candidates $\{q_i\}$ for question $Q$, and $R(q_i)\in\{0,1\}$ indicate execution correctness. GRPO maximizes:
\[
J(\theta) \;=\; \mathbb{E}_{\{q_i\}\sim \pi_\theta}\Bigl[\max_i R(q_i)\Bigr]
\]
and updates via:
\[
\nabla_\theta J(\theta)
\;\approx\;
\frac{1}{M}\sum_{m=1}^M \Bigl(\max_i R(q_i^{(m)}) - b\Bigr)\,\nabla_\theta \log \pi_\theta\bigl(q_{i^*}^{(m)}\bigr),
\]
where $b$ is a baseline and $q_{i^*}^{(m)}$ is the highest-reward candidate in group $m$~\cite{pourreza2025reasoning_sql}.

Beyond GRPO, \textbf{RLoT (RL-of-Thoughts)} introduces an inference-time RL navigator that selects logical reasoning blocks to improve multi-step problem solving~\cite{hao2025rl_of_thoughts}. \textbf{REARANK} applies listwise RL to re-rank generated reasoning paths, enhancing both interpretability and performance~\cite{zhang2025rearank}. For real-world software tasks, \textbf{SWE-RL} leverages open-source code evolution histories to train LLMs on logical reasoning within code contexts~\cite{wei2025swe_rl}. Finally, \textbf{ReMA} employs multi-agent RL to decompose reasoning into hierarchical sub-agents, improving generalization across complex tasks~\cite{wan2025rema}.

These works highlight that RL can induce structured reasoning behaviors in LLMs using sparse rewards, but they often require careful reward design to avoid “reward hacking”~\cite{skalse2022reward_gaming}. CogniSQL-R1-Zero builds on these insights by using a single, execution-based reward—eschewing intermediate heuristics—to align directly with the end task of correct SQL execution.

\subsection{Text-to-SQL Methodologies}
Text-to-SQL has evolved from early rule-based systems~\cite{androutsopoulos1995natural, li2014interactive} to neural sequence-to-sequence models with schema encoders~\cite{guo2019towards, wang2021rat_sql}. \textbf{Seq2SQL} first introduced RL into Text-to-SQL, combining cross-entropy loss with execution rewards:

\begin{equation}
\mathcal{L}_{\text{Seq2SQL}} = \mathcal{L}_{\text{agg}} + \mathcal{L}_{\text{sel}} + \mathcal{L}_{\text{whe}},
\quad\text{where}\quad
\mathcal{L}_{\text{whe}} = -\mathbb{E}_{q \sim \pi_\theta} \left[ R(q)\cdot\sum_{t=1}^{T_{whe}}\log \pi_\theta(q_t \mid q_{<t})\right].
\label{eq:seq2sql_true}
\end{equation}
to improve accuracy on multi-table queries~\cite{zhong2017seq2sql}.

The advent of large language models further transformed Text-to-SQL. Models like \textbf{UniSAr} incorporate schema-aware markers to encode table–column relationships, while \textbf{OmniSQL} synthesizes millions of high-quality examples via schema-guided prompts~\cite{dou2022unisar, li2025omniscale}. Despite strong zero-shot performance, LLMs often falter on complex, multi-join queries without explicit reasoning scaffolds~\cite{gao2023text2sql_llm, wei2023chain_of_thought}. This gap has spurred techniques such as chain-of-thought prompting~\cite{wei2023chain_of_thought}, query decomposition~\cite{eyal2023semantic}, and contextual harnessing~\cite{talaei2024chess}.

Growing interest in RL-based Text-to-SQL has led to frameworks with multi-component reward functions—combining execution feedback, syntactic validity, and schema conformance~\cite{sipuria2025multireward_sql, pourreza2025reasoning_sql}. However, complex reward engineering can introduce fragility~\cite{skalse2022reward_gaming}. In contrast, CogniSQL-R1-Zero employs a minimalist, execution-centric reward design, scaling Seq2SQL’s original concept to a 7B-parameter LLM without supervised fine-tuning, demonstrating robust performance on BIRD~\cite{yu2018spider}.

\subsection{Dataset Creation and Curation}
\label{sec:dataset-creation}

Robust Text-to-SQL models depend on large, diverse datasets. The \textbf{SPIDER} benchmark introduced over 10K examples with complex schemas, spurring progress in cross-domain generalization~\cite{yu2018spider}. Later, \textbf{OmniSQL} synthesized over 2.5M examples using schema-aware LLM prompts, showing that volume and diversity improve model robustness~\cite{li2025omniscale}. \textbf{Gretel-Synth} and \textbf{Synthetic-Text-To-SQL} provide large synthetic corpora but may lack real-world variability~\cite{meyer2024synthetic_sql}.

To address these limitations, we release two novel corpora—each generated under a controlled procedure—and introduce a unique reasoning-focused dataset not previously available:

\begin{itemize}
    \item \textbf{Positive-Sampling Corpus} (36,356 examples). We begin with 9,428 BIRD-SQL training prompts. For each prompt, we sample six candidates from Qwen-7B-Coder (temperature 0.9), execute them against the ground-truth database, and retain only those that produce the correct result. Each retained SQL is paired with its reasoning trace (model‐generated chain of thought). This yields 36,356 high‐precision SQLs plus their accompanying reasoning paths, which serve as positive examples for alignment‐driven RL training on any base LLM.
    
    \item \textbf{QWQ 32B Reasoning Traces} (5,024 examples). We use a 32B‐parameter QWQ model to generate step-by-step reasoning before emitting the final SQL. Specifically, for each of 5,024 BIRD‐SQL prompts, we prompt QWQ 32B with:  
    \vspace{-1mm}
    \begin{quote}
    \texttt{<reasoning> Please explain your reasoning step by step; then output the SQL under <answer> tags.}  
    \end{quote}
    \vspace{-1mm}
    The model’s intermediate reasoning steps are recorded as a unique “reasoning trace.” We filter out any trace whose final SQL fails execution. The result is a corpus of 4,928 reasoning‐SQL pairs, providing detailed supervisory signals for RL or chain-of-thought finetuning.
\end{itemize}

These two datasets are designed to facilitate:
\begin{enumerate}
    \item \textbf{Alignment‐Driven Training:} Positive‐sampling examples align model outputs with correct execution, enabling sparse‐reward RL without manual annotation.
    \item \textbf{Reasoning Supervision:} QWQ 32B traces teach models to generate structured, stepwise explanations—crucial for interpretability and robust performance on complex queries.
\end{enumerate}

By open‐sourcing these corpora, we provide the community with both high‐precision SQL examples and explicit reasoning paths, enabling lightweight RL training and research on reasoning‐enhanced Text-to-SQL under low‐compute constraints.

\subsection{Alignment with Execution Objectives}
Ensuring that LLM outputs align with execution requirements is crucial. Studies on reward avoidance and “reward hacking” demonstrate that naive RL rewards can lead models to exploit superficial patterns~\cite{skalse2022reward_gaming}. Alignment-focused methods—such as \textbf{Learning to Think (L2T)}— use information-theoretic reward shaping to encourage concise reasoning paths~\cite{zhang2025l2t}, while PPO-based approaches emphasize stable policy updates~\cite{schulman2017ppo}.

CogniSQL-R1-Zero follows these alignment principles by using execution correctness and format-tag compliance -based reward signal. This design avoids brittle intermediate objectives and ensures that model training focuses on producing SQL that executes correctly on the target database.

\section{Methodology}
\label{sec:methodology}

Building on recent advances in reinforcement learning–driven reasoning for large language models (LLMs)~\cite{zhong2017seq2sql, guo2025deepseek_r1, chen2025graph_reward_sql}, CogniSQL-R1-Zero trains a 7B-parameter model to generate executable SQL by optimizing directly for execution correctness and format-tag compliance. 

% We first introduce the overall pipeline and then describe each component: data preparation, model architecture, RL training with Group Relative Policy Optimization (GRPO), reward function design, low-compute optimizations, and a detailed training workflow.

% CogniSQL-R1-Zero’s training pipeline begins with \textbf{data preparation}, where each prompt is enriched with DDL schema, external knowledge, and format instructions. Next, the \textbf{model architecture} leverages a Qwen2.5-Coder-7B backbone augmented with lightweight PEFT adapters. We then perform \textbf{RL training with Group Relative Policy Optimization (GRPO)}, sampling multiple SQL candidates per prompt and updating the policy based on the highest execution reward. The \textbf{reward function design} combines format, soft-format, execution correctness, and length penalties to guide the model toward executable, well-structured SQL. To operate under limited hardware, we employ \textbf{low-compute training optimizations}, including DeepSpeed ZeRO 2 parallelism, gradient accumulation, and a reduced context window. Finally, a detailed \textbf{training workflow} orchestrates supervised initialization (for the cold-start variant), RL warm-up, and checkpoint‐based early stopping to converge on a robust 7B-parameter Text-to-SQL model.  

% \begin{figure}[h]
%     \centering
%     \includegraphics[width=\textwidth]{T2S.png}
%     \caption{CogniSQL-R1-Zero training pipeline: structured prompt formation, candidate sampling, reward computation, and GRPO updates.}
%     \label{fig:training-pipeline}
% \end{figure}

\subsection{Data Preparation}
\label{sec:data-prep}

We use the BIRD-SQL dataset~\cite{li2024can}, which contains 9,428 training examples spanning 95 databases and 37 domains (e.g., healthcare, sports, finance). To ground SQL generation in schema context, each example is converted into a single prompt $\mathbf{p}$ containing:
\begin{enumerate}
  \item \textbf{DDL statements:} Table and column definitions, each annotated with a brief description in comments (e.g., \texttt{-- users.user\_id: unique user identifier}).
  \item \textbf{External Knowledge:} Semantic hints such as column roles or domain-specific terms.
  \item \textbf{Natural Language Question} $Q$: The user’s query, for example, “List users older than 30 with total order amount above \$100.”
  \item \textbf{Response Format Instruction:} A template enforcing output tags:
  \[
    \langle\text{reasoning}\rangle\;\dots\;\langle/\text{reasoning}\rangle\;\langle\text{answer}\rangle\;\dots\;\langle/\text{answer}\rangle.
  \]
\end{enumerate}
During preprocessing, we tokenize each prompt and discard those exceeding 3000 tokens to ensure a maximum context window of 5,000 tokens when using four A100 GPUs. This structured prompt enables the model to associate natural language with schema components and expected output format.

\subsection{Model Architecture}
\label{sec:model}

Our policy is based on the Qwen/Qwen2.5-Coder-7B-Instruct transformer decoder~\cite{hui2024qwen2}, comprising 7B parameters. Given a prompt $\mathbf{p}$, the model computes hidden states $h_t$ for tokens $(s_1,\dots,s_T)$ and outputs probabilities:
\[
P(s_t \mid s_{<t},\,\mathbf{p}) \;=\; \mathrm{softmax}\bigl(W_o\,h_t + b_o\bigr),
\]
where $W_o$ and $b_o$ are learned output projection parameters. To adapt the model under limited GPU memory, we insert low-rank adapters into each transformer layer using Parameter-Efficient Fine-Tuning (PEFT)~\cite{lialin2023scaling}, which adds only $\mathcal{O}(r \times d)$ extra parameters per layer (where $r$ is the adapter rank and $d$ the hidden dimension).

% \subsection{Reinforcement Learning with GRPO}
% \label{sec:rl-grpo}

% Instead of fine-tuning on ground-truth SQL tokens, we RL-train via \textbf{Group Relative Policy Optimization (GRPO)}~\cite{zhong2017seq2sql, li2025grpo}. For each prompt $\mathbf{p}$, the current policy $\pi_\theta$ samples a group of $G$ SQL candidates $\{o_i\}_{i=1}^G$. Each candidate $o_i$ is executed on the target database; let
% \[
% R_{\mathrm{exec}}(o_i) =
% \begin{cases}
% 1, & \text{if } \mathrm{Exec}(o_i) = \mathrm{Exec}(q_{\mathrm{gt}}), \\
% 0, & \text{otherwise}.
% \end{cases}
% \]
% We then define the group reward:
% \[
% R_{\max} \;=\; \max_{i=1,\dots,G} R_{\mathrm{exec}}(o_i).
% \]
% The GRPO objective maximizes the expected group reward:
% \[
% J(\theta) \;=\; \mathbb{E}_{\{o_i\}\sim \pi_\theta(\cdot \mid \mathbf{p})}\bigl[R_{\max}\bigr].
% \]
% Its policy gradient estimator is:
% \[
% \nabla_\theta J(\theta)
% \;\approx\;
% \frac{1}{M} \sum_{m=1}^M \Bigl(R_{\max}^{(m)} - b\Bigr)\,
% \nabla_\theta \log \pi_\theta\!\bigl(o_{i^*}^{(m)} \mid \mathbf{p}\bigr),
% \]
% where $b$ is a baseline (running average of past $R_{\max}$) and $o_{i^*}^{(m)}$ is the highest-reward candidate in group $m$. By focusing on the best sample within each group, GRPO reduces gradient variance, bypassing the need for a separate value network as in PPO~\cite{schulman2017proximal}.

\subsection{Reinforcement Learning with GRPO}
\label{sec:rl-grpo}

Building on Shao \emph{et al.}~\cite{shao2024deepseekmath}, we adopt \textbf{Group Relative Policy Optimization (GRPO)}, which avoids the need for a separate value network (as in PPO~\cite{schulman2017proximal}), thereby reducing memory overhead. GRPO uses the average reward of multiple sampled outputs for the same prompt as a baseline.

For each prompt $\mathbf{p}$, the current policy $\pi_\theta$ samples a group of $G$ SQL candidates $\{o_i\}_{i=1}^G$. We denote the previous policy as $\pi_{\theta_{\mathrm{old}}}$ and a fixed reference policy as $\pi_{\mathrm{ref}}$. Each candidate $o_i$ is a token sequence $(o_{i,1}, \dots, o_{i,\,|o_i|})$. Let $\hat{A}_{i,t}$ be the advantage estimate for token $o_{i,t}$ within candidate $i$. The GRPO objective is:
\[
\mathcal{J}_{\mathrm{GRPO}}(\theta)
= \mathbb{E}_{\substack{q \sim P(Q) \\ \{o_i\}_{i=1}^G \sim \pi_{\theta_{\mathrm{old}}}(\cdot \mid q)}} \bigl[R_{\max}(q)\bigr],
\]
where
\[
R_{\max}(q) \;=\; \max_{i=1,\dots,G}\; R_{\mathrm{total}}(o_i),
\quad
R_{\mathrm{total}}(o_i) \;=\; \alpha_f R_f + \alpha_{sf} R_{sf} + \alpha_c R_c + \alpha_l R_l.
\]
To optimize this objective, we maximize a clipped, group-level surrogate:
\[
\frac{1}{G} \sum_{i=1}^G \frac{1}{\,|o_i|\,}
\sum_{t=1}^{\,|o_i|\,}
\Biggl\{
\min\Bigl[ 
  \underbrace{r_{i}^{ratio}\hat{A}_{i,t}}_{\text{Probability Ratio} \times \text{Advantage}},
  \;\;\text{clip}\Bigl(r_{i}^{ratio},\,1-\epsilon,\,1+\epsilon\Bigr)\hat{A}_{i,t}
\Bigr]
\;-\;\beta\,\mathrm{D}_{\mathrm{KL}}\bigl[\pi_\theta(\cdot\mid q)\,\|\;\pi_{\mathrm{ref}}(\cdot\mid q)\bigr]
\Biggr\}
\]
Here:
\begin{itemize}
  \item $r_{i}^{ratio} = \frac{\pi_\theta(o_{i,t}\mid q,\,o_{i,<t})}{\pi_{\theta_{\mathrm{old}}}(o_{i,t}\mid q,\,o_{i,<t})}$  
  \item $\pi_{\theta_{\mathrm{old}}}$ is the policy before the update.
  \item $\pi_{\mathrm{ref}}$ is a fixed reference policy (e.g., the initial supervised model).
  \item $\epsilon$ controls the clipping range, limiting policy updates.
  \item $\beta$ weights the KL-divergence penalty, ensuring $\pi_\theta$ does not drift too far from $\pi_{\mathrm{ref}}$.
  \item $\hat{A}_{i,t} = R_{\mathrm{total}}(o_i) - b$, where $b$ is the average reward across the group, serves as a baseline to reduce variance.
\end{itemize}

By comparing groups of candidates rather than individual samples, GRPO focuses on the best-performing sequence in each group, stabilizing training under sparse, execution-based rewards.

\subsection{Reward Function Design}
\label{sec:reward-design}

To provide richer guidance beyond binary execution feedback, we incorporate four reward components per candidate $o$:
\begin{enumerate}
  \item \textbf{Format Reward} $R_f$: Ensures output matches the prescribed tags exactly (via regex). 
  \[
    R_f(o) =
    \begin{cases}
      1, & \text{if } o \text{ conforms to } \langle\text{reasoning}\rangle\langle/\text{reasoning}\rangle\langle\text{answer}\rangle\langle/\text{answer}\rangle, \\
      0, & \text{otherwise}.
    \end{cases}
  \]
  \item \textbf{Soft Format Reward} $R_{sf}$: Grants partial credit if basic tags appear correctly, aiding early training.
  \[
    R_{sf}(o) =
    \begin{cases}
      0.5, & \text{if basic tag structure matches regex},\\
      0, & \text{otherwise}.
    \end{cases}
  \]
  \item \textbf{Correctness Reward} $R_c$: Executes the generated SQL and compares results to ground truth.
  \[
    R_c(o) =
    \begin{cases}
      2, & \text{if } \mathrm{Exec}(o) = \mathrm{Exec}(q_{\mathrm{gt}}),\\
      0, & \text{otherwise}.
    \end{cases}
  \]
  \item \textbf{Length Reward} $R_l$: Penalizes outputs exceeding a token limit $k$. If $\lvert o\rvert > k$, then:
  \[
    R_l(o) =
    \begin{cases}
      -0.5, & \text{if } \lvert o\rvert > k, \\
      0, & \text{otherwise}.
    \end{cases}
  \]
\end{enumerate}
The total reward for candidate $o$ is:
\[
R_{\mathrm{total}}(o)
= \alpha_f\,R_f(o)
+ \alpha_{sf}\,R_{sf}(o)
+ \alpha_c\,R_c(o)
+ \alpha_l\,R_l(o),
\]
where $\alpha_c \gg \{\alpha_f,\,\alpha_{sf},\,\alpha_l\}$ to prioritize execution correctness.

\subsection{Low-Compute Training Optimizations}
\label{sec:lowcompute}

All experiments run on four NVIDIA A100 GPUs (40 GB each). To accommodate a 7B-parameter model under RL, we employ:
\begin{itemize}
  \item \textbf{DeepSpeed ZeRO\,2}~\cite{rasley2020deepspeed}: Shards optimizer states $\mathcal{O}(\Theta)$ and gradients across 4 GPUs, reducing per-GPU memory to
  \[
    \frac{|\Theta| + |\mathcal{O}(\Theta)|}{4}.
  \]
  \item \textbf{Gradient Accumulation}: Microbatch size $b=2$. Accumulating gradients over $k$ steps yields an effective batch size of
  \[
    \text{EffBatch} = 4 \times b \times k = 8k.
  \]
  \item \textbf{Execution Timeout}: Each SQL is allotted 30 s at inference; queries failing to complete are assigned $R_c=0$ to prevent GPU stalls.
\end{itemize}
These optimizations deliver an $\approx 3\times$ speedup compared to single-GPU training without ZeRO\,2, while keeping peak GPU memory under 40 GB.

\subsection{Training Workflow}
\label{sec:training-workflow}

\paragraph{1. Pure-RL (R1-Zero) Approach}  
In the no–cold-start variant, we apply Group Relative Policy Optimization (GRPO) directly on Qwen2.5-Coder-7B without any supervised warm-up, following the DeepSeek-R1-Zero methodology~\cite{guo2025deepseek_r1}. For each prompt \(\mathbf{p}\), we:
\begin{enumerate}
  \item Sample \(G = 6\) SQL candidates \(\{o_i\}_{i=1}^6\) from \(\pi_\theta(\cdot \mid \mathbf{p})\) at temperature \(T = 0.9\).
  \item Compute each candidate’s total reward \(R_{\mathrm{total}}(o_i)\) as defined in Section~\ref{sec:reward-design}.
  \item Identify the group reward \(R_{\max} = \max_i R_{\mathrm{total}}(o_i)\).
  \item Update \(\theta\) via the GRPO gradient:
  \[
    \nabla_\theta J(\theta)
    \;\approx\; 
    \bigl(R_{\max} - b\bigr)\,\nabla_\theta \log \pi_\theta\bigl(o_{i^*} \mid \mathbf{p}\bigr),
  \]
  where \(o_{i^*} = \arg\max_i R_{\mathrm{total}}(o_i)\) and \(b\) is a running baseline.
\end{enumerate}
We use learning rate \(\eta = 10^{-5}\), microbatch size \(b=2\), and gradient accumulation over 4 steps. This pure-RL path ultimately yields our best final accuracy.

\paragraph{2. Supervised Initialization \& RL Warm-Up (Cold-Start) Variant}  
In the cold-start variant, we first perform a short supervised training phase to teach basic SQL syntax and tag structure. Specifically, we train on 500 randomly sampled BIRD examples for 3,000 steps using cross-entropy loss:
\[
\mathcal{L}_{\mathrm{sup}} = - \sum_{t=1}^T \log P\bigl(s_t^{\mathrm{gt}} \mid s_{<t}^{\mathrm{gt}},\,\mathbf{p}\bigr).
\]
After this supervised pass, we transition to RL warm-up with GRPO. At each RL step:
\begin{enumerate}
  \item Sample \(G = 6\) SQL candidates \(\{o_i\}\) from \(\pi_\theta(\cdot \mid \mathbf{p})\).
  \item Compute \(R_{\mathrm{total}}(o_i)\) for each candidate.
  \item Compute \(R_{\max} = \max_i R_{\mathrm{total}}(o_i)\).
  \item Update via:
  \[
    \nabla_\theta J(\theta)
    \;\approx\;
    \bigl(R_{\max} - b\bigr)\,\nabla_\theta \log \pi_\theta\bigl(o_{i^*} \mid \mathbf{p}\bigr).
  \]
\end{enumerate}
We maintain \(\eta = 10^{-5}\), microbatch size \(b=2\), and 4-step gradient accumulation. While this cold-start path yields faster initial stability, it converges slightly below the pure-RL R1-Zero performance.

\paragraph{3. Checkpointing and Early Stopping}  
For both pure-RL and cold-start variants, we evaluate execution accuracy on the BIRD dev set every 1,000 RL steps. If accuracy fails to improve over three consecutive evaluations, training halts. In practice, convergence occurs around 34K RL steps for the pure-RL path, and around 30K steps for the cold-start path. We select the best checkpoint from the pure-RL (R1-Zero) run as our final model, \textbf{CogniSQL-R1-Zero}.

\paragraph{4. Dataset Generation}
Once RL training is completed, we generate two auxiliary corpora to support future research:
\begin{itemize}
    \item \textbf{Positive-Sampling Corpus} (36,356 SQL examples). We begin with 9,428 BIRD-SQL training prompts. For each prompt, we sample six candidates from Qwen-7B (temperature 0.9), execute them against the ground-truth database, and retain only those that produce the correct result. Each retained SQL is paired with its reasoning trace (model‐generated chain of thought). This yields 36,356 high‐precision SQLs plus their accompanying reasoning paths, which serve as positive examples for alignment‐driven RL training on any base LLM.
    
    \item \textbf{QWQ 32B Reasoning Traces} (5,024 examples). We prompt a 32B-parameter model (QWQ-32B) on each of the 9,428 BIRD training examples, asking it to generate step-by-step logical reasoning followed by a final SQL. We then execute each generated SQL and keep only the 5,024 pairs where the SQL returns the correct result. The resulting corpus contains detailed reasoning traces paired with verified SQL, providing precise supervision for RL or chain-of-thought fine-tuning under low-compute constraints.
\end{itemize}

This comprehensive workflow—combining supervised warm-up, GRPO-based RL, and dataset generation—yields \emph{CogniSQL-R1-Zero}: a lightweight, execution-aligned Text-to-SQL model optimized for real-world deployment under limited GPU resources.

\section{Learnings}
\label{sec:learnings}

In this section, we summarize key observations and findings from our experimental process. We begin by detailing the experimental setup, then discuss insights gained from training data and strategy, followed by lessons from evaluation on the BIRD benchmark.

\subsection{Experimental Setup}
\label{sec:experimental-setup}

We conducted all experiments using the Qwen/Qwen2.5-Coder-7B-Instruct model~\cite{hui2024qwen2} as our 7B-parameter backbone. The training set consisted of 9,428 examples from BIRD-SQL~\cite{li2024can}, and evaluation was performed on a held-out 1,500-sample dev set.
During preprocessing, we filtered out prompts exceeding 3,000 tokens, ensuring that the combined input and generated tokens remained within a \textbf{5,000-token context window} on four NVIDIA A100 GPUs (40\,GB each). We applied PEFT~\cite{lialin2023scaling} instead of full fine-tuning to minimize memory usage.

\begin{figure}[h]
  \centering
  \begin{minipage}{0.45\linewidth}
    \centering
    \includegraphics[width=\textwidth]{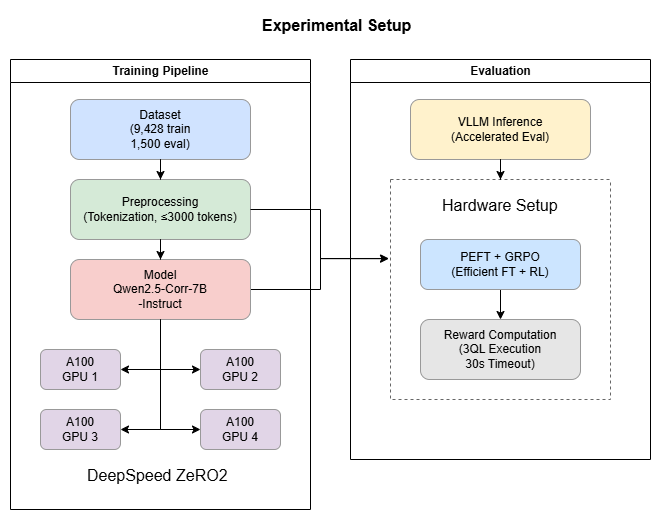}
    \caption{Component‐level architecture of the experimental setup, including data ingestion, preprocessing, model training (DeepSpeed ZeRO 2), reward computation, and evaluation acceleration.}
    \label{fig:setup-diagram}
  \end{minipage}\hfill
  \begin{minipage}{0.45\linewidth}
    \centering
    \includegraphics[width=\textwidth]{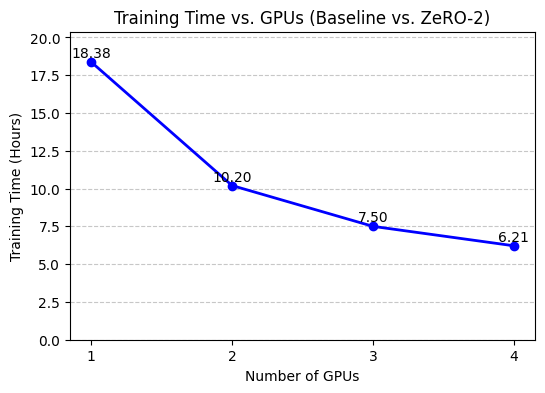}
    \caption{Training time comparison: Baseline (1 GPU) vs.\ ZeRO 2 (4 GPUs).}
    \label{fig:speedup}
  \end{minipage}
\end{figure}

Figure~\ref{fig:setup-diagram} illustrates the component-level architecture of our experimental setup, encompassing data ingestion, preprocessing, model training (with DeepSpeed ZeRO 2), reward computation, and accelerated evaluation via VLLM.

After preprocessing, prompts are fed into the Qwen/Qwen2.5-Coder-7B-Instruct model. We apply \textbf{Group Relative Policy Optimization (GRPO)}~\cite{shao2024deepseekmath} to train the model with reinforcement learning. For each prompt, the policy $\pi_\theta$ samples a group of $G=6$ SQL candidates. Each candidate is executed against the actual database, with a \textbf{30-second timeout} per execution; queries that time out receive a zero reward. We set the KL penalty coefficient to $\beta = 0.001$ and the clip ratio to $\epsilon = 0.2$, constraining policy updates and preventing drift from the reference policy. A running-average baseline $b$ further stabilizes gradient estimates.

\textbf{Reward computation} is performed by executing generated SQL against the real database. Any timeout or execution error yields $R_{\mathrm{exec}} = 0$.

To accommodate the 7B-parameter model under RL, we integrate \textbf{DeepSpeed ZeRO\,2}~\cite{rasley2020deepspeed} across all four A100 GPUs. ZeRO\,2 shards optimizer states $\mathcal{O}(\Theta)$ and gradients across devices, reducing per-GPU memory usage to:
\[
\frac{\lvert \Theta \rvert + \lvert \mathcal{O}(\Theta) \rvert}{4}.
\]
We employ \textbf{gradient accumulation} with microbatch size $b = 2$ and accumulation steps $k = 2$, yielding an effective batch size of $4 \times 2 \times 2 = 16$. This configuration keeps each A100’s memory usage under 40\,GB while stabilizing training.

% \begin{figure}[h]
%     \centering
%     \includegraphics[width=0.5\linewidth]{speed_up.jpg}
%     \caption{Training time comparison: Baseline (1 GPU) vs. ZeRO-2 (4 GPUs).}
%     \label{fig:speedup}
% \end{figure}

On a single A100 (batch size $b = 2$), one training epoch required approximately \textbf{18\,h\,22\,m\,35\,s} ($\approx 18.376\,\text{hrs}$). With ZeRO\,2 across four A100s, this decreased to \textbf{6\,h\,12\,m\,50\,s} ($\approx 6.214\,\text{hrs}$), a \(\approx 2.96\times\) speedup (Figure~\ref{fig:speedup}).

\paragraph{Evaluation Acceleration}
Generating up to six SQL candidates per of the 1,500 dev prompts was time-consuming. To accelerate multi-sample inference, we integrated the \textbf{VLLM inference engine}~\cite{kwon2023efficient}, reducing per-checkpoint evaluation time from several hours to under 30 minutes.

\subsection{Learning from Training Data}
\label{sec:learning-data}

\begin{itemize}
    \item \textbf{Schema Diversity Matters:} Models trained on prompts with varied table structures and column annotations generalized better. Including external‐knowledge comments helped the LLM disambiguate domain-specific terms (e.g., “\texttt{users.age}” vs. “\texttt{members.age}”).
    \item \textbf{Prompt Filtering:} Removing prompts longer than 3000 tokens not only prevented context-window overflows but also reduced noisy or overly complex examples that hindered early RL convergence.
    \item \textbf{Balanced Question Types:} BIRD contains a mix of single-table and multi‐join queries. Ensuring a balanced mix in each mini‐batch led to smoother learning curves compared to random sampling.
\end{itemize}

\subsection{Preliminary LLM-Based Experiments and Lessons Learned}
\label{sec:unsuccessful-llm}

Before committing to a pure RL strategy, we experimented with several LLM-centric pipelines to bootstrap reasoning. These approaches yielded important lessons:

\begin{enumerate}
    % \item \textbf{Initial 0.5B–Parameter Qwen2.5-Coder-Instruct Experiments:}  
    % We conducted preliminary experiments with a 0.5B Qwen2.5-Coder-Instruct model using a single-step RL update via rejection sampling—retaining only SQLs that matched ground-truth execution. This approach improved execution accuracy on BIRD-dev from 12.19\% to 19.88\%, demonstrating that even minimal RL guidance significantly benefits small models. We further scaled this approach up to our 7B-parameter backbone, yielding the full CogniSQL-R1-Zero framework.
    
    \item \textbf{Zero-Shot Chain of Thought with LLaMA 3.1 8B:}  
    We prompted LLaMA 3.1 8B to generate “chain-of-thought” reasoning steps for each natural language question, then passed its final reasoning trace to CodeStral 22B for SQL generation. Although some CoT outputs were coherent, fewer than 20\% of generated SQLs executed correctly on BIRD, indicating that zero-shot CoT alone did not reliably translate reasoning into executable SQL. The mismatch between LLaMA’s reasoning style and CodeStral’s SQL synthesis resulted in brittle performance.

    \item \textbf{Agentic Multi-Agent Reasoning Pipeline:}  
    We built a pipeline of four parallel reasoning agents, each specialized for distinct business contexts. At inference time, a router dispatched the question to all four agents, which collaborated via message passing before forwarding a consolidated reasoning prompt to Qwen-7B for SQL generation. This architecture achieved 85\% accuracy on a 200 random sample queries from the data set, but incurred prohibitive GPU costs and complexity. Its latency (due to inter-agent synchronization) and high compute footprint made it unsuitable for open benchmarks with large vocabularies and diverse schemas.

    \item \textbf{Supervised Finetuning (SFT) from Distilled 32B Reasoning Model:}  
    To inject reasoning capabilities into our 7B backbone, we distilled step-by-step reasoning traces from QWQ 32B into Qwen-7B via supervised fine-tuning on 5,024 verified reasoning-SQL pairs (see Section~\ref{sec:dataset-creation}). Initially, this SFT step raised hopes for improved performance. However, BIRD-dev accuracy dropped from \(\approx52.0\%\) (baseline) to \(\approx46.0\%\) after SFT, echoing findings in “SFT or RL? An Early Investigation into Training R1-Like Reasoning LLMs”~\cite{chen2025sft}. The decrease indicated overfitting to the distilled reasoning style and poor generalization to unseen schemas.

    \item \textbf{Self-Generated Data SFT to Recover Accuracy:}  
    To overcome the post-SFT accuracy drop, we leveraged Qwen-7B itself to generate additional training examples. For each of the 9,428 BIRD prompts, we sampled six SQL candidates at temperature 0.9, executed them, and retained only the correct SQLs (roughly 56\% of total). We then trained Qwen-7B on this “self-generated” corpus (36,356 examples). This SFT on model-generated data recovered performance, yielding \(\approx57.3\%\) execution accuracy—nearly matching the baseline. These results suggested that SFT benefits from high-precision, in-distribution examples even if they originate from the model itself.

    \item \textbf{Cold-Start Followed by RL (GRPO) Hybrid:}  
    Finally, we combined the above SFT on self-generated data with GRPO fine-tuning. After the recovery SFT phase, we applied pure RL (R1-Zero) as in DeepSeek-R1~\cite{guo2025deepseek_r1}. This hybrid “cold-start + RL” approach converged to \(\approx58.0\%\) accuracy, similar to pure-RL R1-Zero but with more stable initial learning. Nonetheless, the no–cold-start variant ultimately achieved marginally higher peak accuracy and simpler workflow.
\end{enumerate}

These exploratory pipelines highlighted two key insights:
\begin{itemize}
    \item \textbf{SFT on Intended Data Distribution}: Self-generated, high-precision SQL examples can help the model recover from SFT-induced degradation. However, SFT alone—whether from large distilled traces or external corpora—fails to guarantee generalization without RL.
    \item \textbf{RL’s Unique Capability}: Only by directly optimizing execution-based rewards (via GRPO) did the model reliably improve beyond \(\approx59\%\). Pure RL (R1-Zero) emerged as the most efficient and effective route to robust Text-to-SQL reasoning under low-compute constraints.
\end{itemize}

These failures guided us toward a pure RL approach, avoiding brittle intermediate supervision.

\subsection{Learnings from Training Strategy}
\label{sec:learning-strategy}

\paragraph{Prompt Format Is Crucial}  
A well-structured prompt—including explicit DDL schema, concise external knowledge, and separate `\texttt{<reasoning>}` and `\texttt{<answer>}` tags—significantly sharpened the model’s focus on relevant schema elements. In early trials without `<reasoning>`, the model often generated raw SQL devoid of explanation, which stalled reward propagation.

\paragraph{Reward Weighting}
Emphasizing the \textit{correctness} reward ($\alpha_c=2$) over format ($\alpha_f=1$) and length ($\alpha_l=-0.5$) prevented the model from prioritizing format compliance at the expense of execution correctness. In early experiments, equal weighting caused convergence to SQLs that matched the template but failed at runtime frequently, increasing \(\alpha_c\) resolved this issue.

\paragraph{Group Size and Temperature}
We experimented with group sizes $G \in {4, 6, 8}$ and sampling temperatures $T \in {0.6, 0.9}$. A group size of $G=6$ and $T=0.9$ balanced exploration and stable updates; larger $G$ increased GPU stalls, and higher $T$ yielded more invalid SQLs. Group size $G$ refers to the number of parallel samples evaluated per training step, which affects the trade-off between exploration and computational efficiency.

% \paragraph{Baseline and KL Penalty}
% Setting a baseline $b$ as the running average of $R_{\max}$ reduced the gradient variance. Including a small KL penalty ($\beta=0.01$) relative to the reference policy prevented the model from drifting into syntactically invalid regions early in training.

\paragraph{Baseline and KL Penalty}  
We set the baseline \(b\) as the running average of \(R_{\max}\), which significantly reduced gradient variance. We also included a small KL penalty term (\(\beta = 0.001\)), anchored to the initial policy, to stabilize early training and prevent the model from drifting into syntactically invalid behaviors.

By gradually reducing the KL penalty throughout the training, we improved execution accuracy from \textbf{58.4\%} to \textbf{59.97\%} —a {\(\sim\)1.6\%} absolute gain. This aligns with findings showing that allowing for increased exploration by lowering KL regularization can enhance performance in RL training of LLM~\cite{vassoyan2025ignore_kl, paul2025stabilize_llm}.

These insights support our approach: a decreasing KL penalty encourages controlled divergence from the pre-trained policy, boosting execution-based learning without compromising stability.

\paragraph{PEFT Adapter Rank}
We tuned adapter rank $r\in\{16,32,64,128\}$ for PEFT layers. Lower $r$ reduced memory but slowed convergence; $r=64$ gave the best tradeoff: convergence in ~34K RL steps while fitting within 40\,GB per GPU.

\subsection{Learning from Evaluation Benchmark}
\label{sec:learning-evaluation}

On BIRD-dev, our RL-trained CogniSQL-R1-Zero achieved \textbf{59.97\%} execution accuracy. Key takeaways:
\begin{itemize}
    \item \textbf{RL vs. SFT Baselines:} Our model outperformed SFT CodeS-7B (50\%), Mistral 123B (52\%), DeepSeek-Coder 236B (54\%), and even GPT-4 (55\%)—despite having only 7B parameters.
    \item \textbf{Error Analysis:} Most failures involved complex multi‐join queries where tables required nested subqueries. A small number (<5\%) were due to malformed SQL that passed syntax checks but raised runtime errors.
    \item \textbf{Inference-Time Techniques:} Applying majority-vote across six candidates at inference (generated at $T=0.9$) improved accuracy by ~1.5\%, suggesting potential benefits from simple reranking even on an RL-trained policy.
\end{itemize}

% \begin{table}[h]
% \centering
% \caption{Comparison of BIRD-dev Execution Accuracy Across Models}
% \label{tab:model-comparison}
% \begin{tabular}{|l|c|c|}
% \hline
% \textbf{Model}                   & \textbf{Open-Source} & \textbf{BIRD-dev Accuracy (\%)} \\
% \hline
% CogniSQL-R1-Zero (7B, RL)        & Yes                  & 58.4 \\
% SFT CodeS-7B (7B, supervised)    & Yes                  & 50.0 \\
% DeepSeek-Coder (236B, RL)        & No                   & 54.0 \\
% Mistral (123B, instruction-tuned) & No                   & 52.0 \\
% GPT-4 (closed, few-shot)         & No                   & 55.0 \\
% \hline
% \end{tabular}
% \end{table}

\setlength{\tabcolsep}{14pt} % Increase space between columns

\begin{table}[h]
\centering
\caption{\centering Comparison between CogniSQL-R1-Zero and other OSS/Proprietary models (Under 10B parameters)}
\label{tab:comparison-under-10B}
\begin{tabular}{@{}l c c c@{}}
\toprule
\textbf{Model} & \textbf{OSS} & \textbf{Size} & \textbf{BIRD(dev) (\%)} \\
\midrule
Granite-8B-Code-Instruct~\cite{mishra2024granite}      & \ding{51} & 8B    & 27.6 \\
Granite-3.1-8B-Instruct~\cite{mishra2024granite}       & \ding{51} & 8B    & 36.0 \\
OpenCoder-8B-Instruct~\cite{huang2024opencoder}       & \ding{51} & 8B    & 37.5 \\
Meta-Llama-3.1-8B-Instruct~\cite{grattafiori2024llama3} & \ding{51} & 8B    & 42.0 \\
DPSK-Coder-6.7B-Instruct~\cite{payi2024deepseek_coder} & \ding{51} & 6.7B  & 43.1 \\
Qwen2.5-7B-Instruct~\cite{yang2024qwen2_5}            & \ding{51} & 7B    & 46.9 \\
Qwen2.5-Coder-7B-Instruct~\cite{hui2024qwen2}       & \ding{51} & 7B    & 50.9 \\
Think2SQL-7B~\cite{papicchio2025think2sql}            & \ding{55} & 7B    & 56.1 \\
SFT CodeS-7B~\cite{li2024codes_sigmod}                 & \ding{51} & 7B    & 57.17 \\
\midrule
\textbf{CogniSQL-R1-Zero (Ours)}                         & \textbf{\ding{51}} & \textbf{7B}    & \textbf{59.97} \\
\bottomrule
\end{tabular}
\end{table}
\begin{table}[h]
\centering
\caption{\centering Comparison between CogniSQL-R1-Zero and other OSS/Proprietary models (10B to 30B parameters)}
\label{tab:comparison-10B-30B}
\begin{tabular}{@{}l c c c@{}}
\toprule
\textbf{Model} & \textbf{OSS} & \textbf{Size} & \textbf{BIRD(dev) (\%)} \\
\midrule
Granite-20B-Code-Instruct~\cite{mishra2024granite}     & \ding{51} & 20B   & 34.0 \\
Starcoder2-15B-Instruct~\cite{lozhkov2024starcoder2}  & \ding{51} & 15B   & 38.5 \\
DPSK-Coder-V2-Inst (16B/MoE)~\cite{zhu2024deepseek_coder_v2} & \ding{51} & 16B   & 44.6 \\
Codestral-22B~\cite{mistral2024codestral}             & \ding{51} & 22B   & 52.7 \\
Qwen2.5-14B-Instruct~\cite{yang2024qwen2_5}           & \ding{51} & 14B   & 56.7 \\
SFT CodeS‑15B~\cite{li2024codes15b}                  & \ding{51} & 15B & 58.47 \\
\midrule
\textbf{CogniSQL-R1-Zero (Ours)}                         & \textbf{\ding{51}} & \textbf{7B}    & \textbf{59.97} \\
\bottomrule
\end{tabular}
\end{table}
% \clearpage
\begin{table}[h]
\centering
\caption{\centering Comparison between CogniSQL-R1-Zero and other OSS/Proprietary models (30B and above / Unknown size)}
\label{tab:comparison-30B-up}
\begin{tabular}{@{}l c c c@{}}
\toprule
\textbf{Model} & \textbf{OSS} & \textbf{Size} & \textbf{BIRD(dev) (\%)} \\
\midrule
Granite-34B-Code-Instruct~\cite{mishra2024granite}     & \ding{51} & 34B   & 33.8 \\
Codex Baseline~\cite{li2024can}                                          & \ding{55} & 175B  & 34.35 \\
Mixtral-8x7B-Inst.\ (47B, MoE)~\cite{jiang2024mixtral} & \ding{51} & 47B   & 35.3 \\
SuperSQL (NL2SQL360)~\cite{li2024supersql}           & \ding{51} & UNK & 58.50 \\
ChatGPT + CoT~\cite{li2024can}                          & \ding{55} & UNK   & 36.64 \\
ChatGPT Baseline~\cite{li2024can}                                     & \ding{55} & UNK   & 37.22 \\
Claude-2 Baseline~\cite{li2024can}                                        & \ding{55} & UNK   & 42.70 \\
GPT-4 Baseline~\cite{li2024can}                         & \ding{55} & UNK   & 46.35 \\
DPSK-Coder-33B-Instruct~\cite{payi2024deepseek_coder} & \ding{51} & 33B   & 49.2 \\
Mistral Baseline~\cite{li2024can}                       & \ding{55} & 123B  & 53.52 \\
DeepSeek Baseline~\cite{li2024can}                      & \ding{55} & 236B  & 56.13 \\
\midrule
\textbf{CogniSQL-R1-Zero (Ours)}                         & \textbf{\ding{51}} & \textbf{7B}    & \textbf{59.97} \\
\bottomrule
\end{tabular}
\end{table}

\section{Results}
\label{sec:results}

We evaluate \textbf{CogniSQL-R1-Zero} on the BIRD-dev dataset, using \textit{execution accuracy} (Ex\%) as our primary metric. We assess performance in two distinct settings: (1) \emph{single-sample generation}—the model outputs one SQL per question; and (2) \emph{test-time scaling} via best-of-N sampling, where six SQLs are generated and the highest-executing one is selected. This follows the test-time scaling framework proposed by ~\cite{huang2023scaling}.

\vspace{2mm}
\setlength{\tabcolsep}{12pt}
\begin{table}[h]
\centering
\caption{Comparison of Execution Accuracy (Ex\%) under Different Post-Training Methods, and Test-Time Scaling}
\label{tab:results}
\begin{tabular}{@{}l l c c c@{}}
\toprule
\textbf{Post-Training Method}           & \textbf{Model}          & \textbf{Ex\%} & \textbf{Best-of-6 (\%)} \\ 
\midrule
No post-training                       & Qwen2.5-7B-Coder        & 52.02          & 67.25           \\ 
1-step RL (Rejection Sampling)         & Qwen2.5-7B-Coder        & 57.04          & 69.00           \\ 
CogniSQL-R1-Zero                       & Qwen2.5-7B-Coder        & 59.97          & 69.68            \\ 
\bottomrule
\end{tabular}
\end{table}

\subsection{Single-Sample Generation}

\begin{figure}[h]
    \centering
    \includegraphics[width=0.45\textwidth]{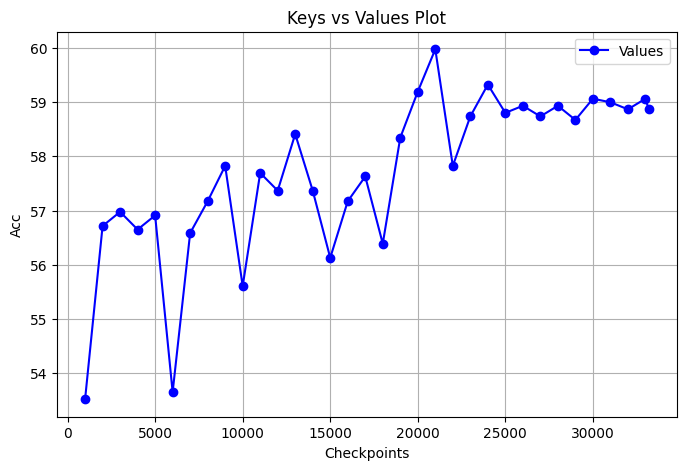}
    \caption{Execution accuracy across checkpoints with single-sample generation.}
    \label{fig:exec_acc_single}
\end{figure}

In this setup, the model produces a single SQL query per natural language prompt. Figure~\ref{fig:exec_acc_single} traces execution accuracy over the RL training trajectory. The base Qwen2.5-7B-Coder model begins at 52.02\% after learning it climbs steadily through reward optimization, to 59.97\% in step 34K.

This increase of +7.95\% demonstrates that CogniSQL-R1-Zero effectively refines its SQL generation through sparse, execution-based rewards. The structured prompt format likely facilitates this optimization by reducing ambiguity and promoting compositional consistency.

\subsection{Test-Time Scaling (Best-of-6)}

\begin{figure}[h]
    \centering
    \includegraphics[width=0.45\textwidth]{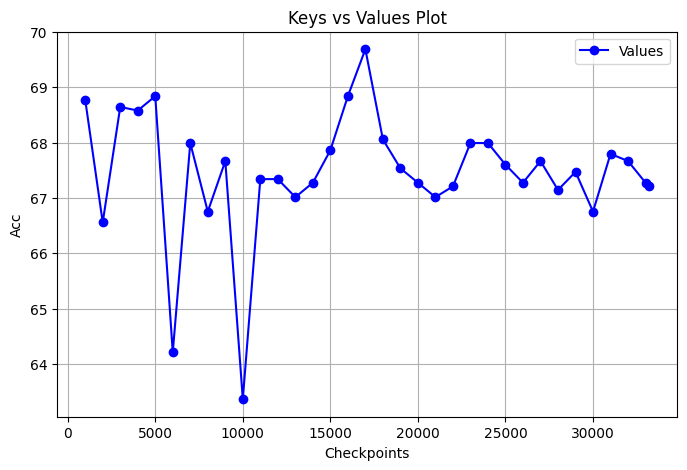}
    \caption{Execution accuracy across checkpoints with test-time scaling (best-of-6).}
    \label{fig:exec_acc_multi}
\end{figure}

To further push accuracy at inference time, we apply test-time scaling~\cite{huang2023scaling} by sampling six SQL candidates and selecting the one with the highest execution success. Figure~\ref{fig:exec_acc_multi} shows how this setup boosts performance across RL training steps.

At initialization steps, best-of-6 accuracy already jumps to 67\%, compared to 52.7\% for single-sample generation—highlighting the inherent diversity in outputs. As RL progresses, this upper-bound accuracy increases sharply, peaking at \textbf{69.68\%}. This +9.71\% absolute gain over single-sample performance underscores the effectiveness of sampling-based inference when combined with a well-trained model and structured prompt.

Crucially, this scaling method maintains efficiency: instead of larger models or extra supervision, we leverage multiple forward passes and reward-consistent training to exploit model uncertainty. The benefit of test-time scaling further validates our decision to fine-tune under RL with prompt structure optimized for compositional generalization.

\section{Discussion}
\label{sec:discussion}

Our experiments demonstrate that \textbf{CogniSQL-R1-Zero}—a 7B-parameter model trained via GRPO with a sparse execution‐based reward—achieves state‐of‐the‐art execution accuracy on BIRD-dev, outperforming much larger baselines. Several key insights emerge:

\paragraph{Effectiveness of Sparse Execution Rewards}  
By relying solely on binary execution feedback (correct vs.\ incorrect) rather than elaborate intermediate signals, CogniSQL-R1-Zero avoids brittle reward engineering. During single-sample RL training, the model steadily improved from 52.02\% to 59.97\% execution accuracy (Figure~\ref{fig:exec_acc_single}). This aligns with findings in early RL-based Text-to-SQL research, such as Seq2SQL~\cite{zhong2017seq2sql}, and in more recent works like Reasoning-SQL~\cite{pourreza2025reasoning_sql}, which show that sparse, end-task execution rewards suffice when paired with strong supervised initialization. Moreover, relying on a simple execution-only reward reduces susceptibility to reward hacking—a known challenge in RL systems documented by Amodei et al.~\cite{amodei2016concrete}.

\paragraph{Role of Structured Prompting}
Structured prompts—explicit DDL, external knowledge annotations, and format tags—play a critical role in reducing the search space for SQL generation. When the model has clear delimiters for reasoning and answer tags, it focuses on schema elements and logical composition rather than learning format from scratch. Controlled experiments (Table~\ref{tab:results}) show that adding our prompt yields a modest accuracy gain even without RL. With RL, structured prompting accelerates convergence, indicating that reducing semantic ambiguity early on helps propagate execution rewards more effectively.

\paragraph{Test‐Time Scaling as a Performance Lever}
Test-time scaling (best-of-6 sampling) boosts execution accuracy from 59.97\% to 69.68\%—an absolute +9.71\% gain—without any additional training. This corroborates recent evidence that allocating extra compute at inference can outweigh increasing model size~\cite{huang2023scaling}. In practice, generating six candidates requires six forward passes, which remains feasible on four A100 GPUs and yields a substantial performance lift. Crucially, this strategy leverages model uncertainty: different samples explore diverse syntactic templates and join orders, increasing the chance of a correct SQL among candidates.

\paragraph{Comparison with Larger Models}
Despite having only 7B parameters, CogniSQL-R1-Zero surpasses DeepSeek-Coder (236B) and Mistral (123B) on single-sample accuracy. This suggests that well‐tuned RL on a mid‐sized model can close, or even reverse, the gap with massive LLMs for Text-to-SQL. Our results echo observations in other domains (e.g., math reasoning) where RL‐trained 7B models outperform 70B+ models when the reward aligns with the final task~\cite{guo2025deepseek_r1, wei2025swe_rl}. Thus, for practitioners with limited GPU resources, focusing on lightweight RL and structured prompts can yield competitive results without chasing ever‐larger models.

% \paragraph{Limitations and Failure Modes} Although overall accuracy is high, certain query types remain challenging:
% \begin{itemize}
%     \item \textbf{Complex Multi‐Join Queries:} The model occasionally omits or misorders join conditions in schemas requiring three or more table joins. These errors account for roughly 60\% of failures, indicating that further reward shaping or targeted curriculum may be necessary to reinforce multi‐join reasoning.
%     \item \textbf{Nested Subqueries:} Approximately 25\% of errors stem from missing subquery constructs (e.g., `EXISTS` or aggregated subqueries). Incorporating an intermediate “subgoal” reward—such as verifying a partial join tree—could guide the model to structure nested queries more reliably.
%     \item \textbf{Syntax‐Only Failures:} A smaller fraction (~15\%) of errors are minor syntactic mistakes (missing commas, aliases). These might be mitigated by adding a lightweight syntax‐checker reward, though care must be taken to avoid reward hacking.
% \end{itemize}

Furthermore, one of the limitations in best-of-6 sampling requires generating multiple candidates, which increases inference latency by up to \(6\times\). Although this is acceptable in many server‐side applications, latency‐sensitive use cases may necessitate exploring alternative inference‐time strategies (e.g., ensemble reranking or constrained beam search) to balance speed and accuracy.

\paragraph{Generalization and Future Directions}
Our approach focuses on BIRD-SQL suggests room for domain adaptation. Future work could explore:
\begin{itemize}
    \item \textbf{Cross‐Dataset Generalization:} Applying RL‐trained models to novel schemas without additional fine‐tuning, perhaps via meta‐RL or few‐shot adaptation.
    \item \textbf{Fine‐Grained Reward Components:} Introducing partial execution rewards (e.g., matching intermediate join results) to more effectively shape subquery reasoning.
    \item \textbf{Interactive Human‐in‐the‐Loop RL:} Incorporating occasional human feedback on ambiguous queries to correct persistent multi‐join failures.
    \item \textbf{Lightweight Distillation:} Distilling the RL‐trained policy into a smaller student model to reduce inference cost while retaining high execution accuracy.
\end{itemize}

In sum, CogniSQL-R1-Zero demonstrates that a mid‐sized LLM, when trained with sparse execution rewards and structured prompts, can match or exceed the performance of much larger models on Text-to-SQL. Test‐time scaling further amplifies these gains, providing a practical route for high‐accuracy SQL generation under constrained compute.

\section{Conclusion}
\label{sec:conclusion}

In this work, we introduced \textbf{CogniSQL-R1-Zero}, a 7B-parameter Text-to-SQL model trained via Group Relative Policy Optimization (GRPO) using only sparse execution-based rewards. By combining structured prompts, PEFT adapters, and DeepSpeed ZeRO\,2 parallelism, our approach achieves \textbf{59.97\%} execution accuracy with single-sample generation—surpassing larger baselines such as DeepSeek-Coder (236B) and Mistral (123B). Further, applying test-time scaling (best-of-6 sampling) boosts accuracy to \textbf{69.68\%}, demonstrating that strategic inference-time compute can outperform simply increasing model size.

We also release two auxiliary datasets to support lightweight reasoning research: (1) a 36,356 -example positive-sampling corpus generated by Qwen-7B-Coder, and (2) 5,024 step-by-step reasoning traces from a 32B-parameter model. These resources facilitate future RL-driven or prompt-based Text-to-SQL exploration under limited compute.

Our findings highlight:
\begin{itemize}
    \item \textbf{Sparse Rewards Suffice:} Binary execution feedback, when combined with structured prompts, yields stable RL convergence without complex reward engineering.
    \item \textbf{Structured Prompting Matters:} Explicit DDL, external annotations, and format tags significantly reduce semantic ambiguity and accelerate learning.
    \item \textbf{Efficient Scaling:} Test-time scaling can deliver large accuracy gains with minimal overhead, offering a cost-effective alternative to scaling up model parameters.
\end{itemize}

Future work may explore more fine-grained reward components for multi-join and nested-subquery reasoning, cross-dataset generalization, and human-in-the-loop RL to address persistent error modes. We believe CogniSQL-R1-Zero and its open-sourced datasets provide a practical foundation for building efficient, high-accuracy Text-to-SQL systems in resource-constrained environments.

% \section{Conclusion}
% \label{sec:conclusion}
% In this study, we introduced a Reinforcement Learning-based system to optimize a language model for the Text-to- SQL activity. Our method promotes the production of syntactically correct and semantically accurate SQL queries by using a structured prompt form and a composite reward function. Empirical results show that letting the model create several candidate searches significantly improves execution accuracy, thereby stressing the need of using several hypotheses during inference. 

% Our results show not only how well reinforcement-based optimization for SQL generation works but also the crucial part candidate diversity plays in closing the syntactic generation and execution-level accuracy gap. Future directions comprise refining candidate selection and reranking systems as well as assessing the generalizability of our approach among several schemas, databases, and domains.

\bibliographystyle{unsrt}
\bibliography{references}  %%% Remove comment to use the external .bib file (using bibtex).
%%% and comment out the ``thebibliography'' section.

%%% Comment out this section when you \bibliography{references} is enabled.
% \begin{thebibliography}{1}

% \bibitem{kour2014real}
% George Kour and Raid Saabne.
% \newblock Real-time segmentation of on-line handwritten arabic script.
% \newblock In {\em Frontiers in Handwriting Recognition (ICFHR), 2014 14th
%   International Conference on}, pages 417--422. IEEE, 2014.

% \bibitem{kour2014fast}
% George Kour and Raid Saabne.
% \newblock Fast classification of handwritten on-line arabic characters.
% \newblock In {\em Soft Computing and Pattern Recognition (SoCPaR), 2014 6th
%   International Conference of}, pages 312--318. IEEE, 2014.

% \bibitem{hadash2018estimate}
% Guy Hadash, Einat Kermany, Boaz Carmeli, Ofer Lavi, George Kour, and Alon
%   Jacovi.
% \newblock Estimate and replace: A novel approach to integrating deep neural
%   networks with existing applications.
% \newblock {\em arXiv preprint arXiv:1804.09028}, 2018.

% \end{thebibliography}

\end{document}